\documentclass[10pt,twocolumn,letterpaper]{article}
\usepackage[accsupp]{axessibility} 
\usepackage{iccv}
\usepackage{times}
\usepackage{epsfig}
\usepackage{graphicx}
\usepackage{amsmath}
\usepackage{amssymb}
\usepackage{dsfont}
\usepackage{multirow}
\usepackage{booktabs} 
\usepackage{siunitx}
\usepackage{colortbl}

\definecolor{mygray}{gray}{.9}

\usepackage[pagebackref=true,breaklinks=true,letterpaper=true,colorlinks,bookmarks=false]{hyperref}

\iccvfinalcopy 

\def\iccvPaperID{2837} 

\ificcvfinal\fi

\begin{document}
\title{Safety-aware Motion Prediction with Unseen Vehicles for Autonomous Driving}
\author{Xuanchi Ren$^{1}$\thanks{Equal contribution} 
\quad  Tao Yang$^2$\footnotemark[1]
\quad  Li Erran Li$^3$ 
\quad  Alexandre Alahi$^4$ 
\quad  Qifeng Chen$^1$\\
$^1$HKUST  \qquad $^2$Xi'an Jiaotong University \qquad $^3$Alexa AI, Amazon \qquad $^4$EPFL\\
}

\maketitle
\ificcvfinal\thispagestyle{empty}\fi

\begin{abstract}
Motion prediction of vehicles is critical but challenging due to the uncertainties in complex environments and the limited visibility caused by occlusions and limited sensor ranges.   In this paper, we study a new task, safety-aware motion prediction with unseen vehicles for autonomous driving. Unlike the existing trajectory prediction task for seen vehicles,  we aim at predicting an occupancy map that indicates the earliest time when each location can be occupied by either seen and unseen vehicles. The ability to predict unseen vehicles is critical for safety in autonomous driving. To tackle this challenging task, we propose a safety-aware deep learning model with three new loss functions to predict the earliest occupancy map. Experiments on the large-scale autonomous driving nuScenes dataset show that our proposed model significantly outperforms the state-of-the-art baselines on the safety-aware motion prediction task. To the best of our knowledge, our approach is the first one that can predict the existence of unseen vehicles in most cases. Project page at {\url{https://github.com/xrenaa/Safety-Aware-Motion-Prediction}}.
\end{abstract}

\section{Introduction}
\label{intro}
Every year, there are more than 1 million deaths related to car accidents, and up to 94\% of accidents are resulted from human errors~\cite{singh2015critical}. Autonomous driving systems can potentially save hundreds of thousands of lives~\cite{blincoe2015economic}. Critical to autonomous driving is motion prediction, which predicts surrounding traffic participants~\cite{zhan2018probabilistic}.

\begin{figure}[t]
\centering
\includegraphics[width=\linewidth]{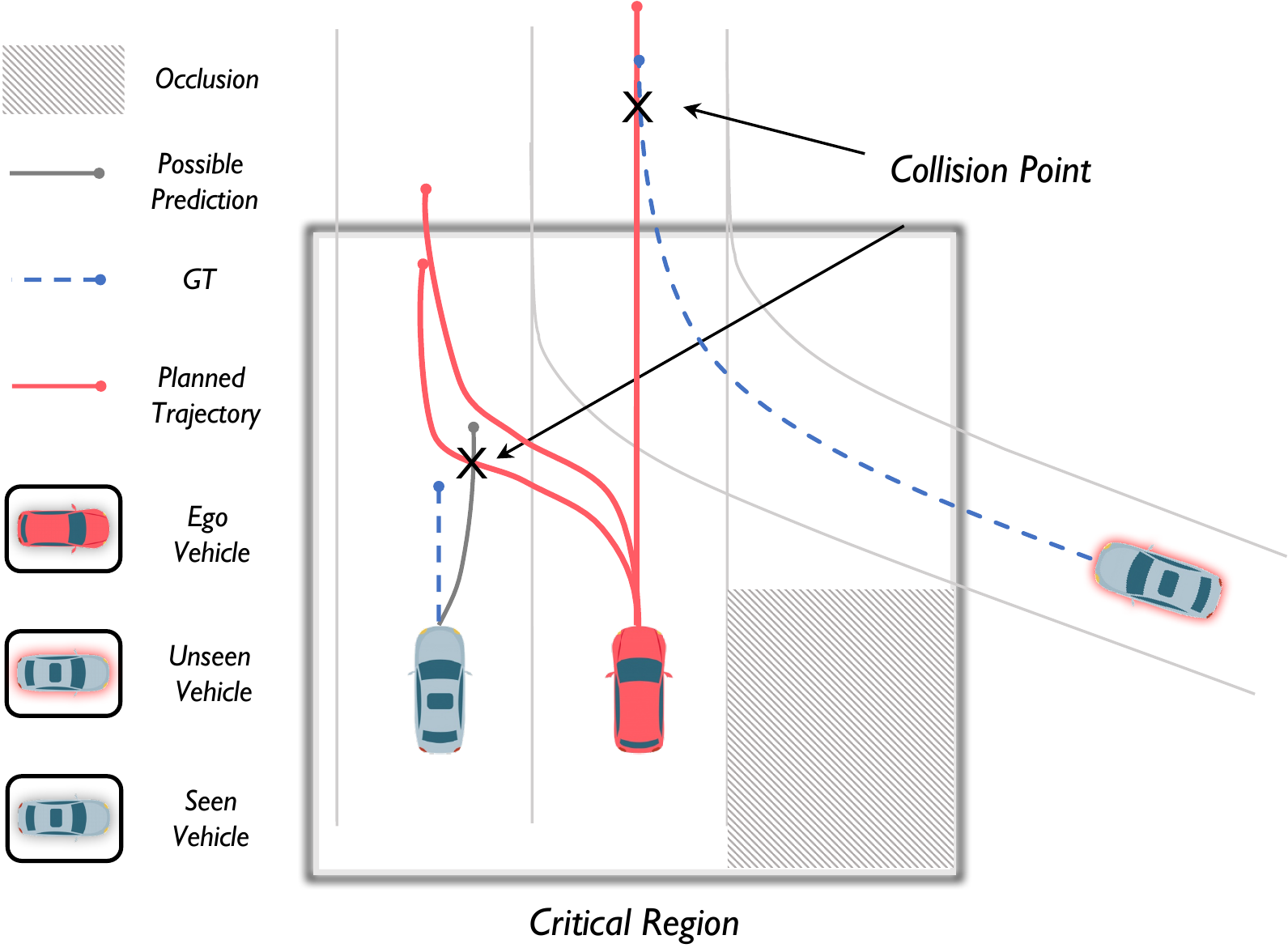}
\caption{
Our goal is to predict how early a vehicle or even unseen vehicles will occupy the space, referred to as \emph{safety-aware motion prediction}.
An unseen vehicle due to the occlusion or limited sensor ranges is the one that can not be observed by the ego vehicle in the past. Ignoring the future motion of unseen vehicles can lead to collisions.
In this figure, the possible prediction (in gray) can help the planner to filter out the risky planned trajectories that may lead to collisions. Safe planning should leave a larger margin for the ego vehicle to respond.
}
\label{fig:demon}
\vspace{-1em}
\end{figure}

Prior work on motion prediction can be broadly classified into two approaches. The first approach predicts the future trajectories of agents. Both discriminative models ~\cite{helbing1995social, wang2007gaussian, lee2016predicting, morton2016analysis, yang2020traffic} and generative models~\cite{zhu2015convex, erlien2013safe, liu2018convex, naumann2019safe, ding2019safe, tacs2018limited} are proposed. The second approach formulates this problem as an occupancy map prediction problem~\cite{hoermann2018dynamic,jain2020discrete,ridel2020scene, liang2020garden, SadatCRWDU20}. These prior work rarely model safety explicitly and have difficulty predicting unseen vehicles.

In real-world driving scenarios, unseen vehicles are very common due to occlusions and the limited range of sensors. An \textit{unseen vehicle} refers to a vehicle that has not appeared at present or in history but will come into view and influence planning decisions. 
An example of an unseen vehicle is illustrated in Figure~\ref{fig:demon}. Missing the prediction of unseen vehicles threatens the safety of planning decisions and even causes collisions.

To achieve safety-first autonomous driving, we analyze the possible consequences of later/earlier prediction, \textit{i.e.}, predicting vehicles' arrival (occupancy) time at a certain location later/earlier than the ground truth, in a specific driving scenario.
As shown in Figure~\ref{fig:demon}, the ground truth (GT) for the surrounding vehicle (blue car) is plotted in a blue dotted line. 
Due to uncertainty, it is hard to make a perfect prediction. In this case, it is safer to make a prediction earlier than the GT, \textit{i.e.}, the predicted arrival/occupancy time at any location is earlier than GT.
When we make an earlier prediction (gray line) than GT, there is a collision with a candidate trajectory. Though the GT trajectory actually does not have a collision with this candidate trajectory, it is safe for the planner to filter out this trajectory. Instead, if the prediction is later than GT, the planner may select a risky candidate trajectory. With the above observations, we propose the task of \textit{safety-aware motion prediction} that includes the following two aspects:
\begin{enumerate}
\vspace{-1mm}
\item For the sake of safety, the predicted occupancy time of each location should be earlier than the ground truth but as accurate as possible.
\vspace{-1mm}
\item 
The prediction for unseen vehicles should be included.
\end{enumerate}

To solve the proposed safety-aware motion prediction task, we propose a new representation called \textbf{earliest occupancy map} to characterize vehicles' future motion (usually in 3 to 5 seconds).
The earliest occupancy map contains a value at each location indicating when this location was first occupied. 
To estimate the earliest occupancy map, we can formulate a regression problem with three novel loss functions. Two of the loss functions encourage accurate prediction with a preference for earlier than later predictions. The third one optimizes for unseen vehicle prediction.
Moreover, with the raster image~\cite{MTP_new} as input and the earliest occupancy map as the output, we propose a new network architecture that uses a customized U-Net~\cite{Unet} with a dilated bottleneck and an unseen-aware self-attention unit. 
Our architecture takes advantage of image-to-image translation networks to model the complex motion prediction task.

Our main contributions are summarized as:
\begin{itemize}
\item
We propose a \textit{safety-aware motion prediction} task for autonomous driving. The task predicts the \textbf{earliest occupancy map} from surrounding vehicles, including both seen and unseen vehicles.
\item
We present a customized U-Net~\cite{Unet} architecture with a dilated bottleneck and an unseen-aware self-attention unit to obtain the earliest occupancy map. Consequently, we introduce three specific loss functions to train our model effectively.
\item 
We introduce new evaluation metrics such as Missing Rate, Aggressiveness, and Unseen Recall to evaluate our models and baselines. The experimental results on the large-scale nuScene dataset show that our model outperforms the state-of-the-art methods for safety-aware motion prediction.
\end{itemize}

\section{Related Work}

\textbf{Motion prediction.}
Accurate motion prediction is critical for autonomous driving~\cite{cosgun2017towards, ziegler2014making}. Deep learning approaches are now state-of-the-art. They have three key components, which are input representation, output representation, and models. 

For the input representations, researchers propose to use either graph-based representations~\cite{homayounfar2019dagmapper, chu2019neural, gao2020vectornet, liang2020learning, TrajectronPlusPlus,kothari2021its,pip,TPCN} or rasterization-based representations~\cite{MTP,bansal2018chauffeurnet,chai2019multipath, HongSP19,bahari2021}.
Homayounfar et al.~\cite{homayounfar2019dagmapper} propose to model the lane graph with a Directed Acyclic Graph (DAG), and Chu et al.~\cite{chu2019neural} use an undirected graph to model the road layout.
Djuric et al.~\cite{MTP_new} rasterize map elements (\textit{e.g.}, roads, crosswalks) as layers and encode the lanes and vehicles with different colors. Compared with graph representation,  raster maps provide richer geometric and semantic information for motion prediction~\cite{liang2020learning}.

For the output representation, prior work has focused on trajectories~\cite{MTP, TrajectronPlusPlus, chai2019multipath} or occupancy maps~\cite{hoermann2018dynamic, jain2020discrete,ridel2020scene, liang2020garden, SadatCRWDU20}. Notably, P3~\cite{SadatCRWDU20} recently propose a semantic occupancy map to enrich the traditional occupancy map~\cite{elfes1989using}.

Prior work leverages either discriminative models~\cite{helbing1995social, wang2007gaussian, lee2016predicting, morton2016analysis, yang2020traffic} or generative models~\cite{trajectron,zhao2019multi,gupta2018social,sadeghian2019sophie,kosaraju2019social,Kothari2021cvpr}. Discriminative models predict either a single most-likely trajectory per agent, usually via supervised regression~\cite{chai2019multipath} or multiple possible trajectories using multi-modal loss function such as mixture-of-experts loss~\cite{MTP}.
Generative models~\cite{TrajectronPlusPlus,tang2019mfp, desire17} explicitly handle multimodality by leveraging latent variable generative models, which incorporate random sampling during training and inference to capture future uncertainty.
However, prior work on motion prediction does not explicitly consider safety and unseen vehicles.
In this paper, we propose the earliest occupancy map as an output representation to assist autonomous driving systems for safety-aware motion prediction with unseen vehicles.

\textbf{Safety and uncertainty awareness.}
Prior work on safety and uncertainty-aware autonomous driving systems has focused on uncertainty estimation~\cite{wolf2008artificial, berthelot2012novel, stellet2015uncertainty, constantin2014margin} and planning with collision avoidance guarantee~\cite{zhu2015convex, erlien2013safe, liu2018convex, naumann2019safe, ding2019safe, tacs2018limited}.
However, it is not straightforward to extend these methods to be unseen vehicles-aware.
To the best of our knowledge, there are few works considering unseen vehicles for the autonomous driving system.
The only exception is Tas and Stiller~\cite{tacs2018limited}, which proposes a method to remain collision-free while considering unseen vehicles during planning. 
However, their method is based on hand-craft rules for each scenario considered (\textit{e.g.} intersection crossing, give-way maneuvers) and can not generalize well to complex urban environments. 

\section{Safety-aware Motion Prediction}
\label{sec:safety}

\begin{figure}[t]
\begin{tabular}{c@{\hspace{0.2em}}c@{\hspace{0.2em}}c}
{\includegraphics[width=0.32\linewidth]{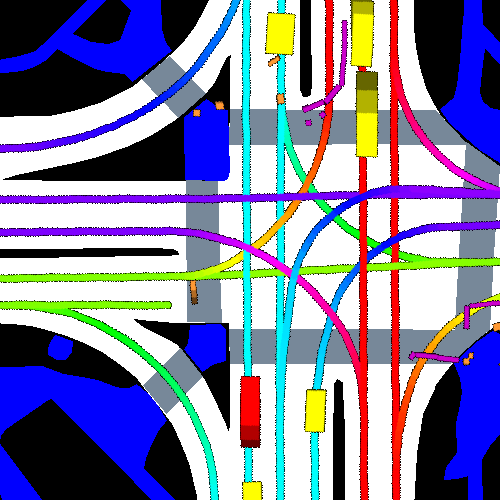}} &
{\includegraphics[width=0.32\linewidth]{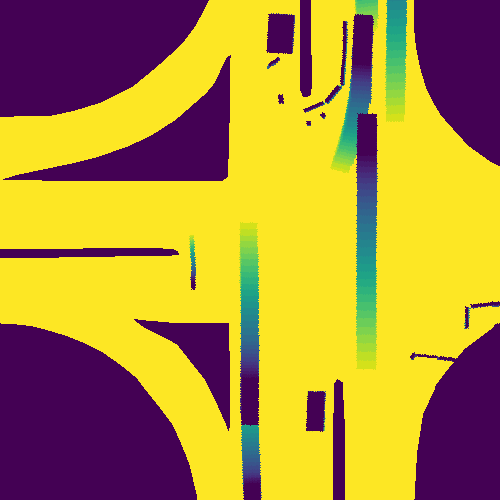}} &
{\includegraphics[width=0.32\linewidth]{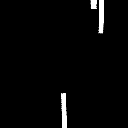}} \\
History & Future & Unseen mask \\
\end{tabular}
\vspace{1mm}
\caption{An example of a scene with unseen vehicles. 
History is represented by the raster image~\cite{MTP_new}.
The ego vehicle is colored red, the other vehicles are colored yellow, and pedestrians are colored orange. Each agent's historical polygons are colored with the same color but with a reduced level of brightness over time. Future is represented by the earliest occupancy map, where the value of each location indicates the earliest time being occupied (darker indicates a smaller value). The unseen mask indicates the locations occupied by unseen vehicles in the future.
}
\label{fig:unseen}
\vspace{-1.5em}
\end{figure}

\subsection{Problem Definition}
\label{subsec:def}
Motion prediction is a necessary component for planning in autonomous driving~\cite{pip}. 
We refer to the area in which motion prediction is needed in order for the planner of the ego vehicle to select a safe trajectory as \emph{critical region}, which is also assumed to be larger than visible-region-with-ego-sensors. 
In an ideal case, the predictions for all the agents in a given scene are needed, such that the critical region is the whole scene.
However, since the receptive ranges of the sensors are limited and occlusions are very common~\cite{yang2020traffic}, we can only assume the critical region to be a neighborhood bounding box of the ego vehicle to simplify the problem.
Under this circumstance, as shown in Figure~\ref{fig:demon}, there is a vehicle that can not be observed by the ego vehicle at present or in history but will enter the critical region in the future and influence the decision of the planner.
Furthermore, as introduced in Section~\ref{intro}, it is also unsafe that prediction is later than ground truth in the real world. 
Therefore, the \emph{safety-aware motion prediction} is defined as predicting the \textbf{earliest occupancy map} that is earlier than the ground truth but as accurate as possible, which also takes the prediction of the unseen vehicles into consideration.
We provide a more thorough problem definition in the supplementary materials.

\subsection{Problem Formulation}
\label{subsec:form}
Given a scene $s$, as assumed in Section \ref{subsec:def}, the critical region $I$ is a neighbourhood bounding box of the ego vehicle to simplify the problem, \textit{i.e.}, $I$ = $\{\,(x,y) \,| l \leq x \leq p,\, m \leq y \leq k,\, x,y\in \mathbb{Z}\}$, where the center position of ego vehicle is $(0,0)$. 
At the current time $t$, considering the historical motion for the previous $H$ timesteps of the agents in the critical region $I$ and the geometric semantic maps of the scene $s$, our target is to predict the future motion of all the agents of the next $T$ timesteps, which also includes the unseen vehicles. The unseen vehicles refer to the vehicles that not in the critical region $I$ at or before time $t$ but enter it in future $T$ timesteps.

\textbf{Occupancy map.}
The occupancy map at time $t$ indicates the occupancy status of each location in the critical region. Let $B_t$ denote the set of the occupied pixels of agents in the scene at time $t$ and $D_t$ denote the pixels of the drivable area. 
We define the occupancy map $O_t$ at time $t$ as follows,
\begin{equation}
\label{equ:occupy_map}
    O_t(x,y) = \left\{
    \begin{aligned}
    1, & \; & (x,y) \in B_t\bigcup \overline{D_t}\\
    0, & \; & \text{otherwise}
    \end{aligned}\right. ,\,\forall (x,y) \in I,
\end{equation}
where ``overline'' indicates the complement. 

\textbf{Earliest occupancy map.}
The earliest occupancy map indicates the timestamp that the position is first occupied. Thus, we formulate the earliest occupancy map $E(x,y)$ as
\begin{equation}
\label{equ:earliest}
    E(x,y) = \min (\{\Delta t|O_{t+\Delta t}(x,y) = 1\}\cup \{T \}),\, \forall (x,y) \in I,
\end{equation}
where $t + \Delta t$ is a timestep between $t$ and $t+T$. Recalling our definition in Section~\ref{subsec:def}, our goal is to derive a prediction $P(x,y)$ that is earlier than the ground truth $E(x,y)$ but as accurate as possible. We formulate it by defining the hard loss (for safety) and the soft loss (for speed). We use the hard loss to penalize predictions later than the ground truth:
\begin{equation}
\label{equ:hard}
    \mathcal{L}_h = \sum_{(x,y)\in I}\mathds{1}(P(x,y) > E(x,y)).
\end{equation}
The hard loss constrains the prediction $P(x,y)$ to be upper bounded by ground truth $E(x,y)$. Only the hard loss will lead to trivial solutions, \textit{i.e.}, all the values are zeros. We add a soft loss to make the prediction close to the ground truth. The soft loss is defined as
\begin{equation}
\label{equ:soft}
    \mathcal{L}_s = -\sum_{(x,y)\in I}P(x,y).
\end{equation}

\textbf{Unseen mask.}
For the prediction of unseen vehicles, we apply an unseen vehicle loss on the predicted earliest occupancy map.
We first introduce an unseen mask to make our model focus on the prediction of unseen vehicles, where the unseen mask covers all the locations occupied by any unseen vehicles in the future, as illustrated in Figure~\ref{fig:unseen}.

With this unseen mask $M$, the unseen vehicle loss is defined as follows,
\begin{equation}
\label{equ:unseen_loss}
    \mathcal{L}_u = \sum_{(x,y)\in I}M(x,y)\mathds{1}((P(x,y) > E(x,y))).
\end{equation}
Note that the losses defined above $\mathcal{L}_h$, $\mathcal{L}_s$ and $\mathcal{L}_u$ are for a single scene $s$.

\begin{figure*}[t]
\centering
\includegraphics[width=\linewidth]{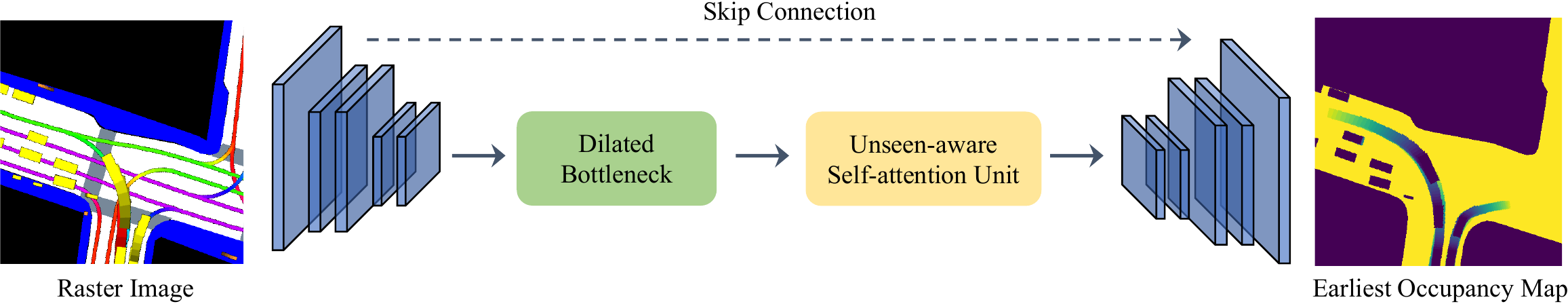}
\caption{Overview of the proposed framework. A raster image is processed by a U-Net to generate the earliest occupancy map. Inside the U-Net, a dilated bottleneck is used to enlarge the receptive field. For the unseen vehicles, we design an unseen-aware self-attention unit.}
\label{fig:overview}
\vspace{-1em}
\end{figure*}

\section{Method}
In this section, we introduce the technical components of our framework as shown in Figure~\ref{fig:overview}.

\subsection{Raster Image}
For the input representation, we use a rasterized image of the bird’s eye view~\cite{MTP_new}, as shown in Figure~\ref{fig:unseen}. 
To discuss this in more detail, the map of a scene $s$ can be represented by a raster map which includes the geometry of the road, drivable area, lane structure and direction of traffic along each lane, locations of sidewalks, and crosswalks. 
The bounding boxes of traffic agents at consecutive timesteps in history are rasterized on top of the map in a color fading effect to form a raster image. 
Furthermore, the raster image is rotated such that the ego vehicle’s heading points up. 
In this work, we raster the critical region, as defined in Section~\ref{subsec:form}, as input. 
Here, we do not use any raw sensor data (\textit{i.e.}, camera, LiDAR, or RaDAR) as an additional input.

\subsection{Dilated Bottleneck}
By taking the raster image as input and the earliest occupancy map as output, the motion prediction task can be modeled as an image-to-image translation problem directly.
Thus, we customized a U-Net~\cite{Unet} to address this problem and learn the joint distribution of the motions of different agents through the translation process.
However, the lowest layer of the conventional U-Net architecture has a 
relatively small receptive field, which limits the network to extract only 
local features, \textit{i.e.}, the model only relies on the part of the critical region to predict the motion of a vehicle, which may lead to collisions.

To enlarge the receptive field and utilize the non-local contextual information, we adopt dilated convolutions~\cite{Dilated, VesalRM19} inside the U-Net. 
Dilated convolutions replace the kernels in the standard convolution layers by sparse kernels with the dilation rate, which defines a spacing between the weights in a kernel.
In this way, with a dilation rate of 2, the receptive field size of a $3\times3$ kernel is equal to that of a $7\times7$ kernel without any increase in complexity.
Thus, in the U-Net architecture, we introduce a dilated bottleneck composed of three dilated convolutions to incorporate local and global contextual information.

\subsection{Unseen-aware Self-Attention Unit}

\begin{figure}[h]
\centering
\includegraphics[width=\linewidth]{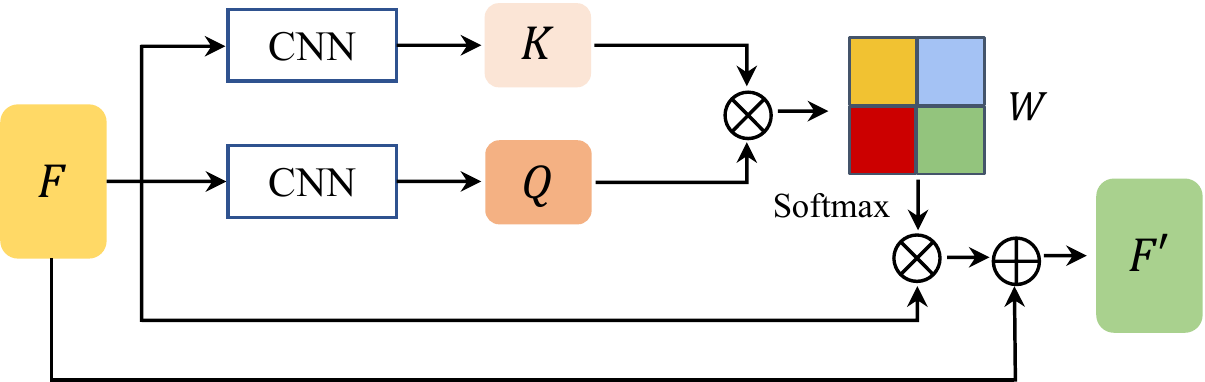}
\vspace{1mm}
\caption{Illustration of the unseen-aware self-attention unit. The input feature map $F$ is fed to two-branch CNNs to generate key $K$ and query $Q$, respectively. Then we put the generated attention mask $W$ on $F$ and use a skip-connection to generate the final output $F'$. $\bigotimes$ denotes element-wise product and $\bigoplus$ denotes element-wise addition. The visualization for the attention mask is shown in Figure~\ref{fig:attention_vis}.}
\label{fig:attention}
\vspace{-1em}
\end{figure}

To make the network focus on unseen vehicles, we design a self-attention unit~\cite{attention} after the dilated bottleneck. Its architecture is presented in Figure~\ref{fig:attention}. The self-attention unit can encode the meaningful spatial importance on feature maps, facilitating the prediction of unseen vehicles.

Given an encoded feature map $F \in \mathbb{R}^{ h \times w \times n}$, where $n$ is the number of channels, $h$ and $w$ indicate the height and width, we feed it into two CNNs respectively to generate the query $Q\in \mathbb{R}^{h \times w \times n}$ and the key $K\in \mathbb{R}^{h \times w \times n}$. Then the attention mask $W$ is defined as
\begin{equation}
    W_{i,j} = \frac{\exp(K_{i,j} \cdot Q_{i,j})}{\sum_{i=1}^{h}\sum_{j=1}^{w}\exp\left(K_{i,j}\cdot Q_{i,j}\right)},
\end{equation}
where $W_{i,j}$ indicates the importance of the feature at $(i,j)$ for predicting unseen vehicles.
In general, we observe that intersection, boundaries and historical occupied region contribute more to the feature map. 
We explore the design of this self-attention unit and empirically find that non-local~\cite{non-local} or only with one CNN branch performs worse than ours, which is presented in supplementary materials.
Additionally, to aggregate the masked feature for unseen vehicles and the original feature, we adopt a skip connection inside the self-attention unit.
Thus, the output $F'$ is finally defined as
\begin{equation}
F' = W \times F + F.
\end{equation}
The final output has both unseen-aware geometric and contextual information and original features, which enhances the performance of the targeting task.

\subsection{Learning}
We train our model in an end-to-end manner. 
Our goal is to make a safety-aware prediction. 
First, we use $\gamma_h \mathcal{L}_h + \mathcal{L}_s$ as one of the optimization objectives, where $\gamma_h$ is a large constant serving as a loss weight. 
As for unseen vehicles, we use $\mathcal{L}_u$ to supervise the learning of unseen vehicles' prediction. 
We thus learn the model parameters by exploiting these loss functions:
\begin{equation}
    \mathcal{L} =\mathcal{L}_{rec} + \gamma_h \mathcal{L}_h + \mathcal{L}_s + \gamma_u \mathcal{L}_u,
\end{equation}
where $\mathcal{L}_h,\, \mathcal{L}_s$ and $\mathcal{L}_u$ are introduced in Section~\ref{sec:safety}. Note that we calculate the average of these losses across all the scenes in dataset $S = \{s_i\}_{i=1}^N$. 
Note that the original equation of $\mathcal{L}_h$ (Eq. \ref{equ:hard}), is not differentiable, we thus use the following equation to approximate it:
\begin{equation}
\label{equ:real_hard}
    \mathcal{L}_h = \sum_{(x,y)\in I} \text{sigmoid} \left( \beta(P(x,y) - E(x,y)) \right),
\end{equation}
where $\beta$ is a large constant. Similarly, we use the same approximation for $\mathcal{L}_u$. 

To stabilize the training, we use the commonly-used pixel-wise mean squared error (MSE) function as the reconstruction term. The reconstruction loss for a single scene is
\begin{equation}
    \mathcal{L}_{rec} = \sum_{(x,y) \in I}\Arrowvert P(x, y) - E(x,y) \Arrowvert^{2}.
\end{equation}

\section{Experiments}

\begin{table*}[t]
\begin{center}
\begin{tabular}{lcccccc}
\toprule
\multicolumn{1}{c}{Method}  & MR (\%) $\downarrow$  & Aggressiveness $\downarrow$  & UR$_{0.3}$ (\%) $\uparrow$ & UR$_{0.5}$  (\%) $\uparrow$ & UR$_{0.7}$  (\%) $\uparrow$ & MSE$\downarrow$\\
\midrule

Physical-CV & 6.53 & 2.77 & 10.98 & 4.58 & 1.46 & 26.26\\
Physical-CA & 6.39 & 2.82 & 11.54 & 4.91 & 1.53 & 26.75\\
Physical-CM & 6.36 & 2.86 & 11.47 & 4.52 & 1.43 & 26.19\\
Physical-CY & 6.48 & 2.81 & 10.95 & 4.39 & 1.43 & 25.89\\
\midrule
MTP & 6.41 & 2.39 & 7.38 & 2.14 & 0.55 & 20.46 \\
Trajectron++ & 8.93 & \textbf{1.71} & 20.40 & 9.33 & 3.45 & 15.38 \\
Trajectron++* & 8.97 & 1.80 & 20.98 & 9.36  & 3.48 & 15.99 \\
P3 & 6.78 & 2.66 & 12.12 & 1.72 & 0.12 & 13.18 \\
\rowcolor{mygray}
Ours & \textbf{1.37} & 2.48 & \textbf{63.28} & \textbf{43.48} & \textbf{18.85} & \textbf{10.61}\\
\bottomrule
\end{tabular}
\end{center}
\caption{Safety-aware motion prediction performance quantitative comparison on the \textbf{nuScenes} dataset. Bold indicates best.}
\label{tbl:quanti}
\end{table*}

\begin{table*}[h]
\begin{center}
\begin{tabular}{lcccccc}
\toprule
\multicolumn{1}{c}{Method}  & MR (\%) $\downarrow$  & Aggressiveness $\downarrow$  & UR$_{0.3}$ (\%) $\uparrow$ & UR$_{0.5}$  (\%) $\uparrow$ & UR$_{0.7}$  (\%) $\uparrow$ & MSE$\downarrow$\\
\midrule
MTP & 8.84 & 5.67 & 51.25 & 8.83 & 0.74 & 77.94 \\
P3 & 9.20 & \textbf{1.48} & 2.76 & 0.24 & 0.00 & 28.54 \\
\rowcolor{mygray}
Ours & \textbf{3.97} & 3.55 & \textbf{88.81} & \textbf{80.22} & \textbf{30.48} & \textbf{18.30} \\
\bottomrule
\end{tabular}
\end{center}
\caption{{Safety-aware motion prediction performance quantitative comparison on the \textbf{Lyft} dataset. Bold indicates best.}}
\label{tbl:lyft}
\vspace{-1em}
\end{table*}

\begin{table*}[t]
\begin{center}
\begin{tabular}{lcccccc}
\toprule
\multicolumn{1}{c}{Method}  & MR (\%) $\downarrow$  & Aggressiveness $\downarrow$  & UR$_{0.3}$ (\%) $\uparrow$ & UR$_{0.5}$  (\%) $\uparrow$ & UR$_{0.7}$  (\%) $\uparrow$ & MSE$\downarrow$\\
\midrule
Ours w/o $\mathcal{L}_h$  & 18.00 & \textbf{1.39} & 36.98 & 20.73 & 6.76 & 6.55 \\
Ours w/o $\mathcal{L}_s$ & 5.85 & 3.53 & 46.31 & 26.36 & 9.91 & 10.76 \\
Ours w/o $\mathcal{L}_u$  & 6.60 & 2.47 & 49.78 & 29.89 & 10.36 & \textbf{6.53} \\
Ours w/o Attention & 5.66 & 2.65 & 39.97 & 22.00 & 6.79 &  6.91\\
Ours & \textbf{1.37} & 2.48 & \textbf{63.28} & \textbf{43.48} & \textbf{18.85} & 10.61 \\
\bottomrule
\end{tabular}
\end{center}
\caption{Ablation study for proposed loss and module.}
\label{tbl:ablation}
\vspace{-1em}
\end{table*}

\subsection{Baselines}
We compare against the following baselines covering different types of methods:

\textbf{Physical models}: Physical models are officially provided by the nuScenes dataset~\cite{nuscenes2019}. There are four settings: 
1) CV: the velocity is constant; 
2) CA: the acceleration and heading are constant;
3) CM: the rates of change of speed and yaw are constant;
4) CY: the (scalar) speed and yaw rate are constant.

\textbf{MTP}: MTP~\cite{MTP} takes the raster image and the target vehicle state as input and predicts the trajectories. 

\textbf{Trajectron++}: Trajectron++~\cite{TrajectronPlusPlus} is a graph-structured recurrent model taking past trajectories of the agents as input to predict the future trajectories. 
Since it does not consider unseen vehicles in the design, 
we also assist Trajectron++ with a handcraft protocol to form a strong baseline: \textbf{Trajectron++*}. We use a Poisson distribution with a hyperparameter $\lambda$ to simulate the number of unseen vehicles during the next $T$ timesteps and put sampled unseen vehicles randomly on the boundary of the critical region at random timesteps.

\textbf{P3}: P3~\cite{SadatCRWDU20} predicts a sequence of occupancy maps from the fused LiDAR and map features. We modify \textbf{P3} to our setting by feeding the raster image as input.

To evaluate the baselines, we convert the outputs of them to the earliest occupancy maps.
For \textbf{Physical models}, \textbf{MTP} and \textbf{Trajectron++}, we fit splines on the predicted trajectories and get the yaw to convert the trajectories to the earliest occupancy map. For \textbf{P3}, we convert the predicted sequence of occupancy maps to the earliest occupancy map, as defined in Eq.~\ref{equ:earliest}.

\subsection{Metrics}

To evaluate the performance of our model, we use a common \textbf{MSE} metric and design three metrics to evaluate safety-aware motion prediction from different aspects. We introduce the metrics below.

\textbf{Missing Rate (MR).} 
For safety, a later prediction is intolerable. Missing Rate indicates the percentage of the predicted earliest occupancy map that is later than the ground truth. 
For $s \in S$, given a predicted earliest occupancy map $P$ and the corresponding ground truth $E$, the Missing Rate can be defined as
\begin{equation}
    {\rm MR} = \frac{\sum_{s\in S}\sum_{(x,y)\in I_s} \mathds{1}(P^s(x,y) > E^s(x,y))} {\sum_{s\in S}|I_s|}.
\end{equation}

\textbf{Aggressiveness.}
The trivial solution for safety-aware motion prediction is that all the cars in the scene $s$ are keeping still. In this case, the values of the earliest occupancy map will be zero. 
However, this is undesirable. Thus, we use the Aggressiveness metric to evaluate if the model has trivial solutions. 
Given predicted earliest occupancy maps $P$, this metric is defined as
\begin{equation}
{\rm Aggressiveness} =  \frac{\sum_{s\in S}\sum_{(x,y)\in \hat{I}_s} (C - P^s(x,y)) } {\sum_{s\in S}|\hat{I}_s|},
\end{equation}
where $\hat{I}_s$ is the subset of $I_s$ containing coordinates that subject to $E^s(x,y) \neq 0$, \textit{i.e.}, $\hat{I}_s = \{(x,y)|(x,y)\in I_s,\, s.t. E^s(x,y)\neq 0\}$, and C is a constant to make the value of the metric positive.

\textbf{Unseen Recall (UR).} 
To evaluate the ability of the model to capture unseen vehicles, we choose to calculate the recall for the prediction of unseen vehicles.
Given the unseen mask $M$, the set of occupied positions of the unseen vehicles is $\hat{M} = \{(x,y)|M(x,y)=1\}$, the IoU for unseen vehicles is defined as
\begin{equation}
{\rm IoU} = \frac{ |\hat{M} \cap  \hat{P}|  }{ |\hat{M}| } ,
\end{equation}
where $\hat{P}$ is the set of positions of the predicted motions, \textit{i.e.}, $\hat{P} = \{(x,y)|t < P(x,y) < t+T\} $.
Then Unseen Recall (UR) is defined as:
\begin{equation}
{\rm UR_{\alpha}} = \frac{ \sum_{s \in \hat{S}} \mathds{1}( \text{IoU}_s > \alpha) }{ |\hat{S}|} ,
\end{equation}
where $\hat{S}$ are the subset of $S$ that containing unseen vehicles. Here, we consider the threshold $\alpha$ to be $0.3$, $0.5$ and $0.7$.

\textbf{MSE.} 
To evaluate the performance of motion prediction models, Average Displacement Error (ADE)~\cite{gupta2018social} is commonly used.
Due to the output occupancy map of our method is image-level, we use the MSE metric between the prediction and GT instead to evaluate the accuracy of the predictions. Note that MSE is only used as a reference.

\subsection{Implementation details}

Our model is implemented in Pytorch~\cite{pytorch} and trained on an NVIDIA V100 GPU in around 24 hours. 
We used a batch size of 32 and trained with Adam optimizer~\cite{adam} with a learning rate to \num{1e-4}. 

The critical region used in our work is of size $50$ meters by $50$ meters. The range in front of the ego vehicle is $40$ meters, and the range at the back is $10$ meters. The ranges for the left and the right are the same, which are both $25$ meters. The pixel resolution of the raster image is 1:10. Thus, $m = -100, k = 400, l = -250, p = 250.$
For all the models, we only feed the agents inside the critical region at present and in history. For the Poisson distribution using the handcrafted protocol, we use a $\lambda = 2$.

For the hyper-parameters, we use the information from the past $2$ seconds to predict the future $3$ seconds. Thus, with a frequency of $10$ Hz, the total number of the historical timesteps $H$ is $20$, and the future timesteps $T$ is $30$. The data provided by the nuScenes dataset is $2$ Hz, so we interpolate it to $10$ Hz~\footnote{For the input raster image, we only use $2$ Hz data.} to make the earliest occupancy map smooth. Considering $T = 30$, we set $C = 31$. For the loss functions, we set $\beta = 100$ and $\gamma_h = \gamma_u = 1000$. The dilated rates used in the dilated bottleneck are $2$, $4$, and $8$, respectively.

\subsection{Evaluation on the nuScenes dataset}

We evaluate our method on the public nuScenes dataset~\cite{nuscenes2019}. 
It is a large-scale dataset for autonomous driving with $1000$ scenes in Boston and Singapore. Each scene is annotated at $2$ Hz and is $20$s long, containing up to $23$ semantic object classes, as well as high definition maps with $11$ annotated layers. 
We follow the official benchmark for the nuScenes prediction challenge to split the dataset. There are $32,186$ prediction scenes in the training set and $8,560$ scenes in the validation set. Due to the inaccessible ground truth of the test set, we use the validation set to evaluate the models for safety-aware prediction.

To understand the prevalence of unseen vehicles,
we compute the number of scenes with unseen vehicles in the nuScenes dataset when the critical region is limited to be $50$ meters by $50$ meters. There are about 47\% of scenes containing unseen vehicles in the training set and about 32\% of scenes in the validation set. 
This indicates that unseen vehicles are common in real-world scenarios.

\textbf{Quantitative Comparison.}
We perform the quantitative comparison on the baselines and our model in terms of the above four metrics. 
Since the baselines do not consider unseen vehicles, to illustrate the effectiveness of our method, we modify Trajectron++ to Trajectron++* with a handcrafted unseen vehicle prediction protocol.
The results are summarized in Table~\ref{tbl:quanti}.
By modeling safety-aware prediction explicitly and using the earliest occupancy map as representation, our model outperforms the state-of-the-art models and traditional physical models except on Aggressiveness. 
However, note that the Aggressiveness metric evaluates if the models have trivial solutions. Therefore, it is not an essential metric for safety-aware prediction.
We can observe that our model has the minimum MR and MSE, which indicates that our model has the fewest cases that prediction is later than ground truth and conforms to the definition for safety-aware prediction: earlier but as accurate as possible.
UR measures if the models can predict the unseen vehicles without omission. Our model achieves the highest recall across different thresholds, indicating that it can predict unseen vehicles effectively.
Note that a handcrafted protocol can not help a lot for unseen vehicle prediction, demonstrating that the prediction for unseen vehicles should take contextual information into consideration.
Furthermore, the deep learning-based baselines do not outperform physical-based methods significantly. 

\begin{figure}[t]
\begin{tabular}{c@{\hspace{0.2em}}c@{\hspace{0.2em}}c}
{\includegraphics[width=0.32\linewidth]{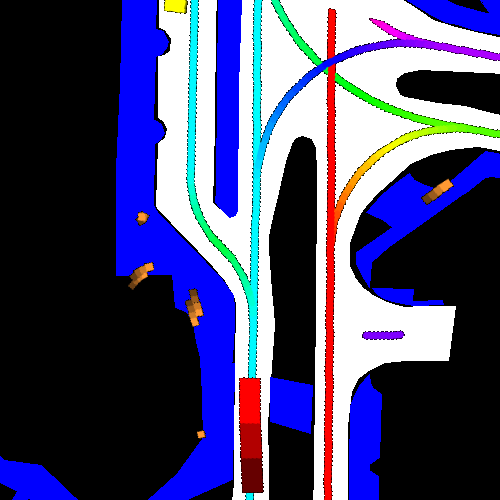}} &
{\includegraphics[width=0.32\linewidth]{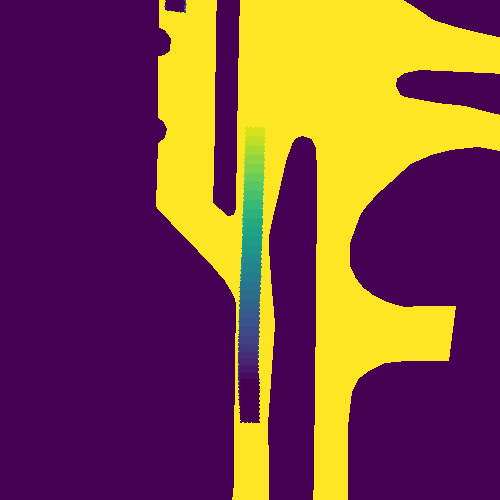}} &
{\includegraphics[width=0.32\linewidth]{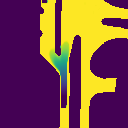}} \\
{\includegraphics[width=0.32\linewidth]{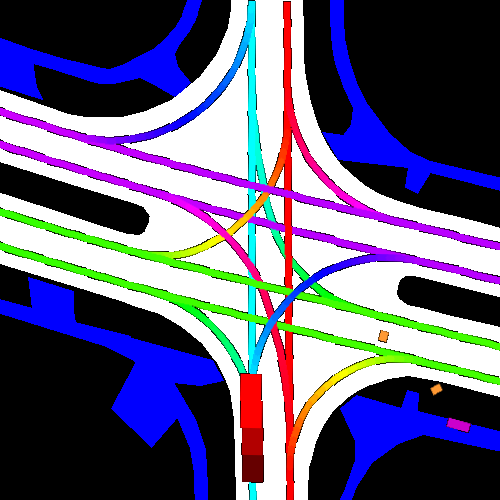}} &
{\includegraphics[width=0.32\linewidth]{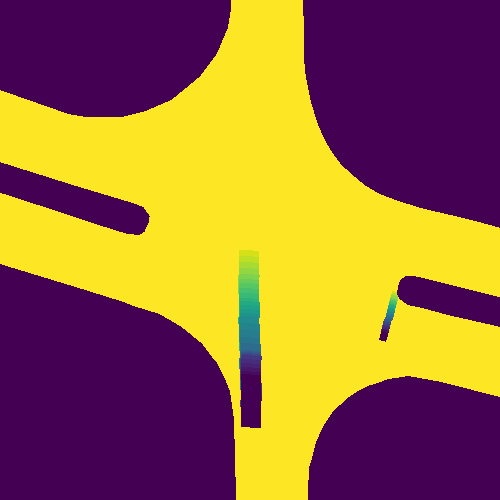}} &
{\includegraphics[width=0.32\linewidth]{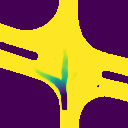}} \\
Input & GT & Ours \\
\end{tabular}
\caption{Multi-modal predictions made by our method. Using the earliest occupancy map, we can achieve multi-modal future predictions without taking an explicit probabilistic approach.}
\label{fig:multi}
\vspace{-1em}
\end{figure}

\begin{figure*}[t]
\begin{tabular}{@{}c@{\hspace{1mm}}c@{\hspace{1mm}}c@{\hspace{1mm}}c@{\hspace{1mm}}c}

{\includegraphics[width=0.192\linewidth]{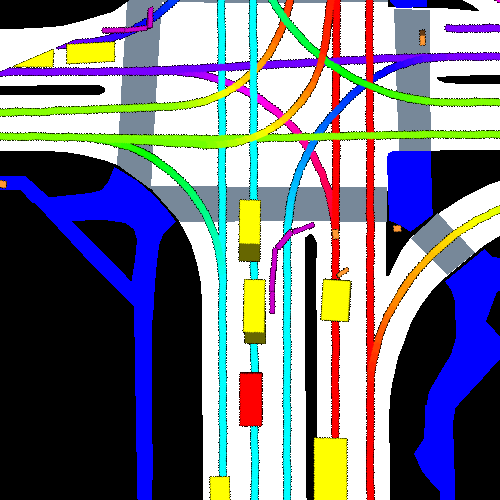}} &
{\includegraphics[width=0.192\linewidth]{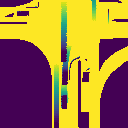}} &
{\includegraphics[width=0.192\linewidth]{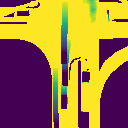}} &
{\includegraphics[width=0.192\linewidth]{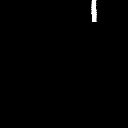}} &
{\includegraphics[width=0.192\linewidth]{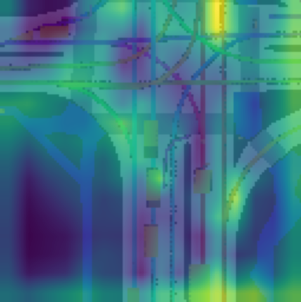}} \\

Input & GT & Ours & Unseen mask & Attention map \\
\end{tabular}
\caption{Visualization of attention masks $W$. For the attention map, the brighter area in the image indicates more significant responses. The attention map has higher responses around the unseen mask, thereby helping the model predict the motion of unseen vehicles.}
\label{fig:attention_vis}
\vspace{-0.5em}
\end{figure*}

\begin{figure*}[t]
\centering
\begin{tabular}{@{}c@{\hspace{1mm}}c@{\hspace{1mm}}c@{\hspace{1mm}}c@{\hspace{1mm}}c@{\hspace{1mm}}c}
\includegraphics[width=0.16\linewidth]{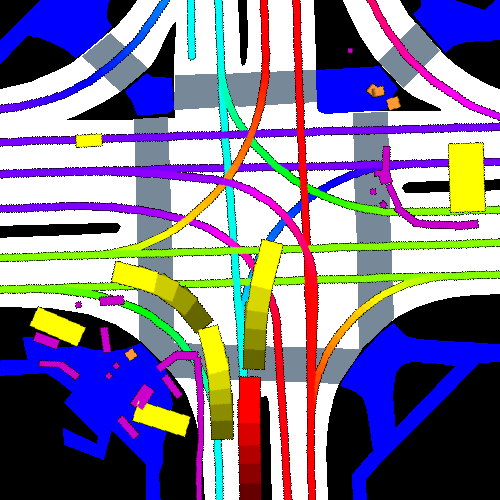} &
\includegraphics[width=0.16\linewidth]{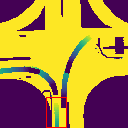} &
\includegraphics[width=0.16\linewidth]{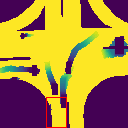} &
\includegraphics[width=0.16\linewidth]{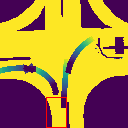} &
\includegraphics[width=0.16\linewidth]{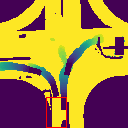} &
\includegraphics[width=0.16\linewidth]{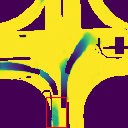}\\

\includegraphics[width=0.16\linewidth]{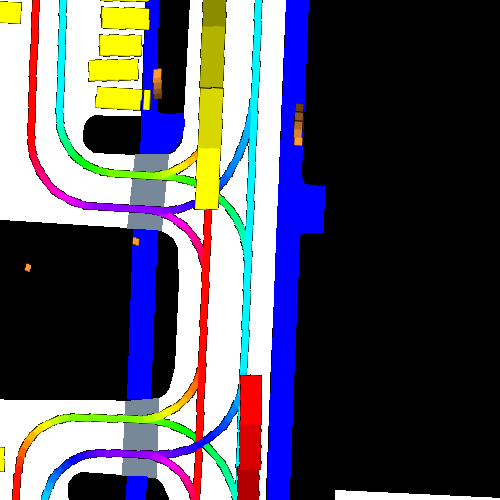} &
\includegraphics[width=0.16\linewidth]{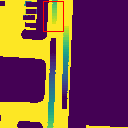} &
\includegraphics[width=0.16\linewidth]{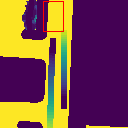} &
\includegraphics[width=0.16\linewidth]{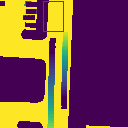} &
\includegraphics[width=0.16\linewidth]{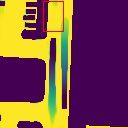} &
\includegraphics[width=0.16\linewidth]{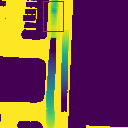}\\
Input& GT & MTP & Trajectron++ & P3 & Ours\\
\end{tabular}
\vspace{1mm}
\caption{Visual comparisons between ours and other baselines on the nuScenes dataset. All the prediction results are visualized with the earliest occupancy maps. The unseen vehicles are annotated with red bounding boxes.
We show common failures of the baselines, including missing predictions for unseen vehicles and later predictions for seen vehicles. Such unsafe predictions could mislead the ego vehicle to make poor planning decisions.
In contrast, the future motion predicted by our method is earlier but as accurate as possible and includes unseen vehicles.
}
\label{fig:visual}
\vspace{-1em}
\end{figure*}

\textbf{Qualitative Comparison.}
We show our prediction results under diverse traffic scenarios and provide some representative comparisons with the deep learning-based baselines in Figure~\ref{fig:visual}. Please refer to supplementary materials for more results.

\textbf{Visualization for attention map.}
We formulate the motion prediction problem as an image-to-image translation problem and train our model with input-output image pairs. Therefore, the prediction for unseen vehicles relies on the data distribution, especially the density of agents and entrance locations. To understand the mechanism of the unseen-aware self-attention unit, we visualize the spatial attention maps by overlaying them on the input images in Figure~\ref{fig:attention_vis}. We can observe that spatial attention helps our model locate the unseen vehicles and drivable regions, which indicates that our model learns the patterns of the data, particularly for the unseen vehicles.

\textbf{Multi-modality.}
Multi-modality gain popularity recently in motion prediction. Instead of using probabilistic approaches, we provide an alternative way by using the earliest occupancy map. As shown in Figure~\ref{fig:multi}, with our proposed formulation and loss, the earliest occupancy map is capable of representing multi-modal predicted motion in a single output. Furthermore, as shown in Figure~\ref{fig:visual}, the motions predicted by our method tend to have a bit larger range and makes the system safety-awareness. For future work, a hierarchical probabilistic U-Net~\cite{multi-Unet} may further improve the ability for multi-modal prediction.

\subsection{Evaluation on the Lyft Dataset}

In this section, we further evaluate our model on the Lyft dataset. 
The Lyft dataset~\cite{lyft} has over $1,000$ hours of driving data in Palo Alto, California. 
It consists of $170,000$ scenes, each of which is $25$ seconds long. It also provides a high-definition semantic map with $15,242$ labeled elements. We follow the official guidelines of the Kaggle challenge to split the dataset. There are $4,009,980$ prediction scenes in the training set.  For the validation, we use $20,000$ scenes, a subset of the official validation set.

Because of the lack of support for the Lyft dataset of the state-of-the-art methods, we only select MTP~\cite{MTP} and P3~\cite{SadatCRWDU20} as the baselines to compare in terms of the above four metrics. The results are summarized in Table~\ref{tbl:lyft}. Compared to baselines, our method achieves the best MR, UR, and MSE, which shows that our model predictions are safe and accurate. 
Our method to detect unseen vehicles relies on the current frame; however, there are many missing frames for agents in the Lyft dataset, which results in more detected unseen vehicles.

\subsection{Ablation studies}

To develop an understanding of which model component influences the performance, we conduct ablation studies on the proposed losses and attention module on the nuScenes dataset. The results are summarized in Table~\ref{tbl:ablation}. 
We have three key observations.
1) The hard loss is essential for safety-aware prediction.  The first row shows that training our model without hard loss results in a significant drop in the MR and UR.
2) The unseen loss, soft loss, and unseen-aware self-attention unit are necessary components for our model. Lacking any one of them hurts the performance in terms of MR and UR.
3) As one would expect, the unseen-aware self-attention unit is more important for learning the prediction of unseen vehicles than other components. Even without the unseen loss as supervision, our method can still outperform the baselines for predicting unseen vehicles.

Note that, without hard loss, the soft loss is adequately optimized. Therefore, ``Ours w/o hard'' achieves the lowest aggressiveness, but it does not mean that it is safer than others. 
The slight drop in the MSE of the final model is due to the wrong prediction of unseen vehicles (false positive).
Overall, the hard loss and unseen-aware self-attention unit are the dominant performance-improving components.

\section{Conclusion}
In this paper, we study a new task named safety-aware motion prediction for autonomous driving. 
The proposed task requires the predicted event (arrival time at a location) to be earlier than the actual event in the future while as accurate as possible. We introduce a novel safety-aware representation called the earliest occupancy map that characterizes the vehicles' future motion.
With this representation, we formulate the safety-aware motion prediction as an image-to-image translation problem. To solve the problem, we present a customized U-Net architecture with a dilated bottleneck to enlarge the receptive field and an unseen-aware self-attention unit to facilitate the prediction of unseen vehicles.
Our model is trained effectively with three novel loss functions.
Experimental results on a large-scale autonomous-driving dataset show that the proposed framework significantly outperforms state-of-the-art baselines on the safety-aware motion prediction task. As for the limitation, our method may have some false positive predictions for the unseen vehicles. Though the false positives do not compromise the safety, they may introduce more constraints for the planner.

\newpage

{\small
\bibliographystyle{ieee_fullname}
\bibliography{egbib}
}

\end{document}